\documentclass{article}

\usepackage{microtype}
\usepackage{graphicx}
\usepackage{subcaption}
\usepackage{booktabs}
\usepackage{hyperref}

\usepackage[preprint]{icml2026}

\usepackage{amsmath}
\usepackage{amssymb}
\usepackage{mathtools}
\usepackage{algorithm}
\usepackage[capitalize,noabbrev]{cleveref}
\usepackage{multirow}
\usepackage{makecell}
\usepackage{paralist}
\usepackage{pdflscape}
\usepackage{threeparttable}
\usepackage{siunitx}
\usepackage{xcolor}

\icmltitlerunning{AgenticSum: Agentic Inference-Time Framework for Faithful Clinical Summarization}

\begin{document}

\twocolumn[
  \icmltitle{AgenticSum: An Agentic Inference-Time Framework for Faithful Clinical Text Summarization}

  \begin{icmlauthorlist}
    \icmlauthor{Fahmida Liza Piya}{ud}
    \icmlauthor{Rahmatollah Beheshti}{ud}
  \end{icmlauthorlist}

  \icmlaffiliation{ud}{University of Delaware, Newark, DE, USA}

  \icmlcorrespondingauthor{Fahmida Liza Piya}{lizapiya@udel.edu}

  \icmlkeywords{Clinical NLP, Hallucination Detection, Agentic AI, Factuality, Large Language Models}

  \vskip 0.3in
]

\printAffiliationsAndNotice{}
\begin{abstract}
Large language models (LLMs) offer substantial promise for automating clinical text summarization, yet maintaining factual consistency remains challenging due to the length, noise, and heterogeneity of clinical documentation. We present \texttt{AgenticSum}, an inference-time, agentic framework that separates context selection, generation, verification, and targeted correction to reduce hallucinated content. The framework decomposes summarization into coordinated stages that compress task-relevant context, generate an initial draft, identify weakly supported spans using internal attention grounding signals, and selectively revise flagged content under supervisory control.
We evaluate \texttt{AgenticSum} on two public datasets, using reference-based metrics, LLM-as-a-judge assessment, and human evaluation. Across various measures, \texttt{AgenticSum} demonstrates consistent improvements compared to vanilla LLMs and other strong baselines. 
% LLM-based evaluation shows reduced hallucination frequency (approximately 10\% on MIMIC-IV and 4\% on SOAP summaries) while preserving completeness and factual consistency. Human evaluation further confirms reduced hallucination severity, with annotator agreement exceeding 78\% across domains. 
Our results indicate that structured, agentic design with targeted correction offers an effective inference-time solution to improve clinical note summarization using LLMs.
\end{abstract}

\paragraph*{Data and Code Availability}
This study uses the MIMIC-IV-Ext-BHC dataset \citep{aali2024mimic} available on PhysioNet and the SOAP summary dataset \citep{neupane2024soap} available on Hugging Face Datasets. Code to reproduce our experiments is available at \url{https://github.com/healthylaife/CHIL_AgenticSUM}.

\paragraph*{Institutional Review Board (IRB)} This study involved human subjects as part of an evaluation survey and was deemed exempt from full Institutional Review Board (IRB) review due to minimal risk.

\section{Introduction}
\label{sec:intro}
Hallucination remains a persistent challenge in clinical text summarization~\citep{wan2024acueval,chuang2024lookback}, reflecting a mismatch between the generative behavior of large language models and the evidentiary and accountability requirements of clinical documentation~\citep{slobodkin2024attribution,goodman2024ai,piya2025advancing}. In this context, (faithfulness) hallucinations typically appear as fluent but unsupported statements, including inferred diagnoses, treatments, or patient attributes that are not explicitly grounded in the source record. Unlike benign errors in open-domain generation, such deviations can misrepresent clinical facts, undermine downstream decision-making, and erode clinician trust. As a result, clinical documentation workflows rely on structured oversight in which draft summaries are reviewed, corrected, and authorized by a responsible practitioner prior to release~\citep{saadat2025enhancing,goodman2024ai}. By contrast, most LLM-based summarization pipelines operate in a single forward pass~\citep{ren2025malei}, with factuality assessed, if at all, only through post hoc evaluation~\citep{grolleau2025medfacteval,croxford2025evaluating} or correction mechanisms~\citep{wan2024acueval}, rather than explicit verification or supervisory control during generation~\citep{ji2023survey}.

Despite their strong generative capabilities, most LLM-based clinical summarization systems generate summaries directly from long, noisy, and redundant clinical notes without explicit mechanisms for verification, correction, or controlled termination~\citep{piya2025contextual}. Clinical documentation often contains templated text, copied sections, fragmented timelines, and institution-specific shorthand, which makes reliable single-pass generation difficult~\citep{rajkomar2018scalable,johnson2023mimiciv}. As a result, unsupported content can propagate unchecked into the final output, a failure mode repeatedly observed in evaluations of LLM-generated medical summaries~\citep{asgari2025framework,joseph2024factpico}. In the absence of explicit supervision or revision control, these systems lack a principled mechanism to determine whether a summary should be corrected, partially revised, or released as-is, even when hallucinations are detected~\citep{mckenna2023sources,huang2025survey}. This stands in contrast to clinical documentation workflows, in which draft summaries are generally reviewed and authorized by a responsible practitioner prior to release~\citep{meligonis2025junior,goodman2024ai}. Consequently, hallucination in clinical summarization reflects not only a modeling limitation but also a mismatch between single-step generation and high-stakes documentation practices.

To address the lack of verification and control in single-pass generation, prior work has explored agentic AI frameworks that introduce structure, feedback, and iterative refinement into language model generation~\citep{shinn2023reflexion,fu2023improving,ke2024enhancing}. By decomposing generation into coordinated stages, these approaches have shown promise in complex reasoning tasks and have attracted interest in clinical and biomedical settings~\citep{asgari2025framework,kim2025medical}. However, most existing agentic systems rely on self-generated critique without explicit grounding in the source document, lack principled supervision over revision termination, and do not align with clinical documentation workflows that require accountable review prior to release~\citep{shinn2023reflexion,madaan2023selfrefine,yao2023react}. Consequently, agentic decomposition alone remains insufficient for faithful and supervised clinical summarization.

While agentic frameworks introduce procedural structure into generation, they do not provide reliable signals for grounding or supervision on their own. In parallel, recent work has explored model-internal signals for assessing factual consistency in generated text~\citep{piya2025advancing}. Decoder-side signals such as attention patterns offer lightweight, token-level indicators of factual grounding without external supervision or fine-tuning~\citep{chuang2024lookback,bui2025correctness,jiang2025hicd}. Considered independently, agentic architectures support iterative refinement, while model-internal signals provide localized diagnostics for factual consistency. However, these approaches remain largely decoupled: agentic systems lack grounded signals to guide revision and termination, and signal-based methods are not integrated into generation or release control~\citep{slobodkin2024attribute}. Consequently, a cohesive framework that unifies procedural oversight with model-internal evidence remains missing.

In this work, we introduce a structured agentic framework for clinical text summarization that explicitly separates generation from verification and revision, enabling systematic hallucination detection and control. The framework decomposes summarization into modular stages for reconstructed input with important context, draft generation, factuality assessment, and targeted correction, yielding a verifiable multi-stage process with interpretable signals of content grounding. Our contributions are as follows.
\begin{itemize}
    \item We introduce \texttt{AgenticSum}, an agent-based clinical summarization framework that decomposes input selection, summary generation, factuality assessment, and targeted correction into distinct, auditable stages. This separation enables structured factual assessment and controlled correction during inference, without fine-tuning or task-specific supervision.
    \item We propose two core mechanisms within the \texttt{AgenticSum} framework: \texttt{FOCUS}, an attention-guided input compression module that prioritizes clinically relevant context during summary draft generation, and \texttt{AURA}, a grounding analysis module that leverages model-internal signals, including decoder attention, to quantify token-level reliance on source content. Together, these components support fine-grained traceability and localized correction of unsupported statements.
    \item We evaluate \texttt{AgenticSum} on MIMIC-IV discharge summaries and SOAP notes, demonstrating consistent improvements in factual accuracy and error localization over single-pass summarization baselines while maintaining comparable fluency under deployment-realistic constraints.
\end{itemize}

\begin{figure*}[htbp]
    \centering
    \includegraphics[width=\textwidth]{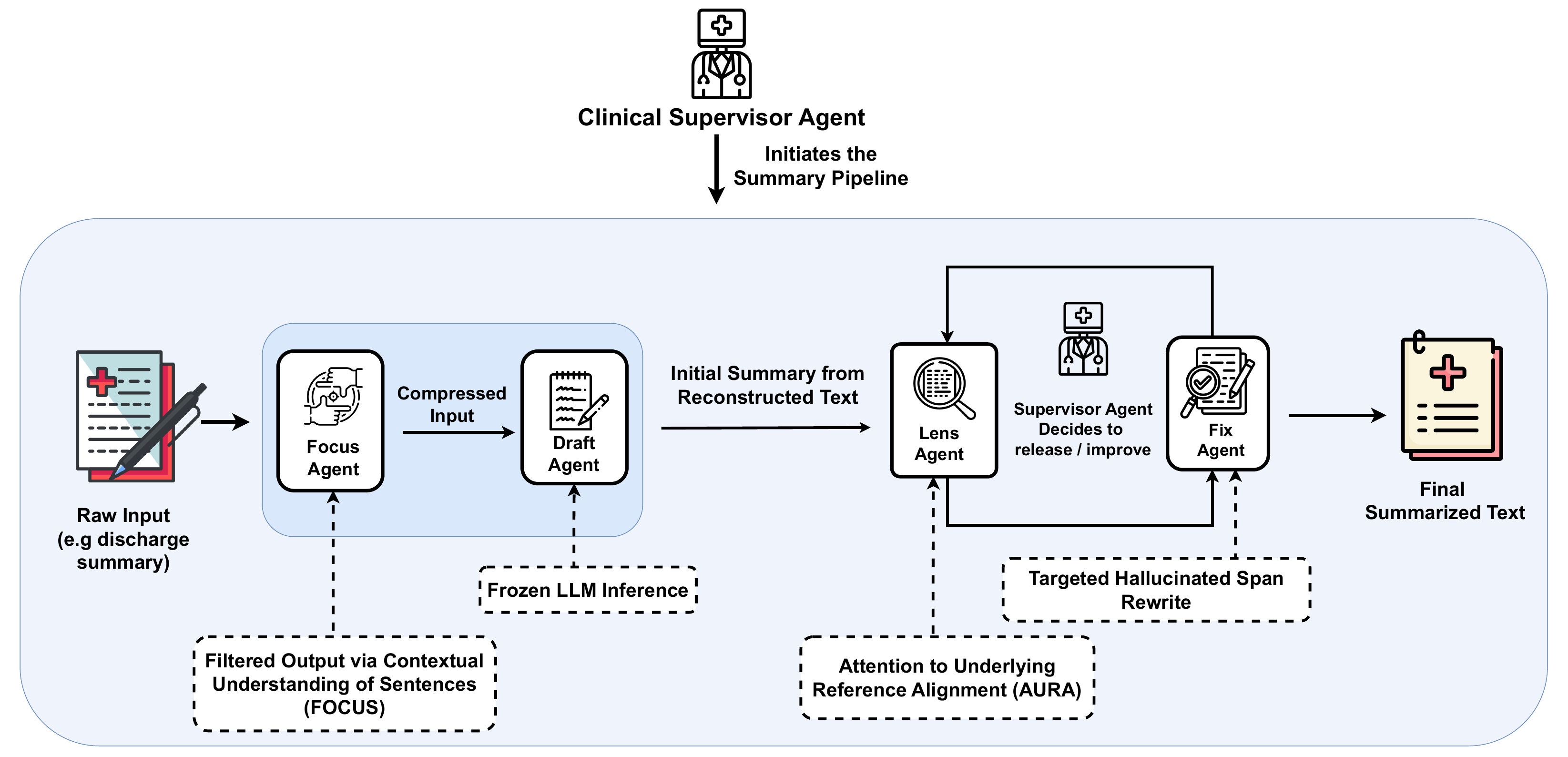}
    \caption{Our agentic  framework, \texttt{AgenticSum}, designed for hallucination-aware clinical text summarization.}
    \label{fig:Method}
\end{figure*}

\section{Related Work}
\label{sec:lit}
Our proposed method draws on several lines of prior work, which we briefly cover below.

\paragraph{Clinical Text Summarization}
Recent advances in clinical summarization have leveraged pre-trained language models~\citep{piya2025advancing}, including domain-adapted models such as \texttt{BioGPT}~\citep{luo2022biogpt} and \texttt{PubMedBERT}~\citep{gu2021domain}, as well as instruction-tuned models like \texttt{Flan-T5}~\citep{lyu2024automatic}. Task-specific architectures have also been proposed~\citep{han2024optimal}, such as \texttt{Pointer-GPT}, to improve content retention. However, most existing approaches operate over uncompressed token sequences, leading to inefficiencies and increased hallucination risk in long clinical narratives. As clinical documents grow longer and more heterogeneous, expanded context length further increases computational cost and exacerbates factual inconsistency, making it harder for models to distinguish source-supported content from prior-driven inferences.

\paragraph{Extrinsic Hallucination, Prior-Driven Generation, and Faithfulness}
In clinical summarization, large language models may produce fluent and clinically plausible content that is not explicitly supported by the source document, a phenomenon broadly characterized as hallucination and particularly concerning in high-stakes settings~\citep{asgari2025framework,huang2025survey}. Many existing summarization models are designed to condense overall records or narrative history, prioritizing coverage and coherence rather than strict source faithfulness. When applied to clinical notes, which require precise attribution to documented evidence,
this design emphasis can lead to unsupported inferences. One well-studied form of hallucination is \textit{extrinsic hallucination}, in which generated statements appear clinically reasonable but are not grounded in the input text~\citep{huang2025survey,mckenna2023sources}. Prior work shows that such errors are often driven by reliance on learned statistical regularities and training-time correlations, especially when inputs are underspecified
or ambiguous. This prior-driven behavior can shift generation away from source-based evidence toward inference, undermining faithfulness even when relevant information is present in the document~\citep{maynez2020faithfulness,pagnoni2021understanding}. As a result, recent evaluation frameworks emphasize faithfulness through source-grounded factuality, assessing whether generated statements are explicitly supported by the input text~\citep{kryscinski2020evaluating,huang2025survey}. Several recent approaches adopt this perspective by leveraging model-internal signals to estimate reliance on source content during generation~\citep{chuang2024lookback,bui2024correctness,jiang2025hicd}.

\paragraph{Token-Level Filtering and Attention-Based Selection}
Prior work has explored improving LLM efficiency through model-level compression techniques such as pruning~\citep{frantar2023sparsegpt}, quantization~\citep{dettmers2022gpt3}, and knowledge distillation~\citep{hinton2015distilling, piya2024healthgat}, reducing memory footprint and inference latency under hardware constraints. These methods operate on model parameters and representations and are largely agnostic to input structure and informational density~\citep{piya2022bdfn}. In clinical NLP, this limitation is critical: aggressive compression can degrade performance on domain-specific tasks requiring fine-grained semantic distinctions, as even low-magnitude parameters may encode clinically salient associations~\citep{tinn2023fine,ma2023llm}. Although distillation transfers general linguistic competence~\citep{ho2022large}, distilled models often fail to preserve nuanced biomedical context and long-range dependencies essential for clinical summarization~\citep{magister2022teaching}.
To address redundancy at the input level, token-level filtering methods reduce effective context length at inference time by selectively retaining informative subsets of the input. Early approaches such as \texttt{PoWER-BERT} prune tokens based on attention scores~\citep{goyal2020power}, but are limited to encoder-based architectures. Subsequent methods introduce differentiable top-$k$ selection with stochastic perturbations, enabling gradient-based optimization at the cost of variance undesirable in safety-critical settings. More recent attention-based techniques retain salient tokens to reduce latency and memory overhead~\citep{lin2024rho,lou2024sparser}, and deterministic, architecture-compatible filtering has been demonstrated for long clinical inputs~\citep{piya2025contextual}. However, token-level selection methods primarily emphasize efficiency and do not define mechanisms for enforcing factual grounding or preventing unsupported content during generation.

\paragraph{Agentic Decomposition for Faithful Clinical Summarization}
Beyond input efficiency, prior work suggests that hallucination in clinical summarization arises not only from insufficient context or inefficient processing, but from generation pipelines that fail to separate generation from verification and correction~\citep{maynez2020faithfulness,dziri2022faithfulness}. In single-pass prompting, generation, factual assessment, and revision are entangled within a single inference step, leaving no principled mechanism to identify or correct unsupported statements once produced. This limitation is particularly problematic in clinical settings, where locally plausible errors can be clinically misleading. These concerns have motivated growing interest in agentic and multi-stage frameworks that decouple generation, evaluation, and revision during inference~\citep{madaan2023selfrefine,shinn2023reflexion}. Prior work shows that separating reasoning and action~\citep{yao2023react}, enabling iterative self-critique~\citep{shinn2023reflexion,madaan2023selfrefine}, and coordinating role-specialized agents can improve robustness and controllability in complex generation tasks. Related studies demonstrate the effectiveness of agentic collaboration for document-level processing, including collaborative transformation and summarization~\citep{fang2025collaborative,li2025chatcite}, and agent-based systems have been explored to support clinical workflows~\citep{yang2025ai}. However, existing approaches primarily emphasize task accuracy, interaction quality, or efficiency, with limited focus on source-grounded faithfulness or explicit hallucination diagnosis in document-level clinical summarization. Consequently, while agentic decomposition offers a promising structural foundation, its integration with factual verification remains unresolved.

\section{Methodology}
\label{sec:method}
We introduce a modular framework for clinical text summarization that explicitly separates generation, verification, and targeted repair to detect and reduce hallucinations. The framework decomposes summarization into role-specialized components for, i) input compression, ii) draft generation, iii) hallucination detection, and iv) constrained revision, implemented by the \emph{FocusAgent}, \emph{DraftAgent}, \emph{HallucinationDetectorAgent}, and \emph{FixAgent}, respectively. Refinement is governed by a supervisory control module, the \emph{ClinicalSupervisorAgent}, which determines whether further correction is required or whether the summary has stabilized and can be finalized. This design enables structured inference-time refinement with explicit termination control, rather than relying on unconstrained self-correction. Figure~\ref{fig:Method} provides an overview of the framework.

\paragraph{Problem Formulation.}
Let \( \mathcal{D} \) denote the space of clinical documents, and let \( D \in \mathcal{D} \) be an input document represented as a sequence of tokens,
\[
D = (t_1, t_2, \dots, t_n),
\]
where \(t_i\) denotes the \(i\)-th input token and \(n\) is the document length. The objective is to generate a summary
\[
S = (y_1, y_2, \dots, y_{L_S}),
\]
that is concise, clinically informative, and faithful to the source document, where \(y_\ell\) denotes the \(\ell\)-th output token and \(L_S\) may vary across documents.

We define hallucination as the generation of statements in \(S\) that are not supported by evidence in the source document \(D\). Rather than optimizing an explicitly constrained objective, we approximate faithful summarization through an inference-time decomposition that separates generation, verification, and revision. Conceptually, the goal is to produce the summary $S^*$ that is likely to be generated under the pretrained language model, while remaining well supported by the source document:
\[
S^* \approx \arg\max_{S} P(S \mid D;\theta),
\]
where \(\theta\) denotes fixed, pretrained model parameters. This objective is approximated through an iterative inference procedure that identifies and revises unsupported spans using explicit grounding and entailment signals derived from \(D\). We define an entailment scoring function \( E(\cdot, \cdot) \), where \( E(s, D) \in [0,1] \) measures the
degree to which a summary span \( s \subset S \) is supported by and consistent
with the content of the source document \( D \), as determined by a pretrained
language model operating in an evaluative role.

%The framework addresses hallucination through a supervised, multi-stage inference process in which content selection, generation, verification, and correction are handled by distinct components rather than being entangled within a single decoding pass. Long clinical documents often contain redundancy and weakly relevant content that can propagate noise into downstream generation and verification. To mitigate these failure modes, summarization proceeds through the following stages:

%\begin{itemize}
%\item \textbf{Input compression.} The \emph{FocusAgent} performs sentence-level input compression using an attention-based mechanism referred to as \texttt{FOCUS}, identifying context-relevant sentences while reducing redundancy.
%\item \textbf{Draft generation.} The \emph{DraftAgent} generates an initial summary \( S \) conditioned on the compressed input.
%\item \textbf{Verification.} The \emph{HallucinationDetectorAgent} evaluates summary spans for evidential support using semantic entailment predictions and an attention-based grounding metric, \texttt{AURA}.
%\item \textbf{Targeted correction.} The \emph{FixAgent} revises only weakly supported spans, preserving supported content and avoiding full regeneration.
%\item \textbf{Supervisory control.} The \emph{ClinicalSupervisorAgent} determines whether additional detect-and-repair cycles are required or whether the summary has stabilized and can be finalized.
%\end{itemize}

\subsection*{FocusAgent: Input Compression via FOCUS}
Clinical narratives often interleave clinically salient information with templated documentation, conversational filler, and tangential discourse, particularly in longitudinal records and patient–provider communications~\citep{chen2025leveraging}. Importantly, conversational segments are not inherently extraneous and frequently contain clinically meaningful information, including symptom descriptions, treatment adherence updates, and patient-reported outcomes that must be preserved during summarization~\citep{chen2025leveraging,piya2025advancing}. Because relevance is often localized and weakly signaled in such settings, relying on generative attention alone conflates evidence identification with text synthesis. We therefore introduce an explicit input compression stage that precedes generation.

We formalize this design as \emph{Filtered Output via Contextual Understanding of Sentences (FOCUS)}, a sentence-level input compression mechanism that selectively retains context salient for summarization. FOCUS is a deterministic, attention-based filtering procedure that leverages decoder self-attention from a frozen transformer model to estimate sentence-level contextual salience. Sentences are scored independently, ranked by salience, and a fixed proportion of the highest-ranked sentences is retained as compressed input for subsequent agents.

Let the input document \(D\) be partitioned into \(m\) sentences,
\[
D = \bigcup_{j=1}^{m} u_j,
\]
where each sentence \(u_j\) corresponds to a contiguous subset of tokens with index set \(T_j \subseteq \{1,\dots,n\}\), and the sets \(\{T_j\}_{j=1}^m\) form a disjoint partition of the document tokens.

FocusAgent performs sentence-level input compression by estimating sentence importance using decoder self-attention from a pretrained transformer model. Each sentence is processed independently in a single forward pass with attention outputs enabled, and the resulting decoder self-attention weights are used solely as an internal signal to estimate contextual relevance rather than for autoregressive generation. Operating exclusively at the sentence level preserves semantic coherence, negation structure, and clinical interpretability while providing a lightweight and deterministic compression mechanism without additional trainable parameters.
Let \(A\) denote the decoder self-attention tensor from the final transformer layer for a given sentence, with shape \(H \times T \times T\), where \(H\) is the number of attention heads and \(T\) is the number of tokens in the sentence. Sentence salience is computed as a single scalar by aggregating decoder self-attention across heads and token interactions, with normalization by sentence length. Formally, the importance score of sentence \(u_j\) is defined as,
\[
\beta_j =
\frac{1}{H T}
\sum_{h=1}^{H}
\sum_{i=1}^{T}
\sum_{k=1}^{T}
A^{(j)}_{h,i,k},
\]
where \(A^{(j)}\) denotes the decoder self-attention tensor corresponding to sentence \(u_j\). Let \(m\) denote the total number of sentences in the input document. FOCUS retains the top \(k\) sentences, where \(k\) denotes the number of sentences selected after compression and is defined as,
\[
k = \lfloor r \cdot m \rfloor,
\]
for a user-selected retention ratio \(r \in (0,1]\). The resulting compressed input is given by
\[
D_{\text{reduced}} =
\{\, u_j \mid \beta_j \ge \beta_{k} \,\},
\]
where \(u_j\) denotes the \(j\)-th sentence in the input document and
\(\beta_{k}\) denotes the \(k\)-th highest sentence salience score.

\subsection*{DraftAgent: Summary Generation}
Given the compressed input \( D_{\text{reduced}} \), the DraftAgent generates an
initial summary using a pretrained decoder-only language model. The compressed
document serves as the sole conditioning context for generation, reflecting the
input selection performed by the FocusAgent.

Formally, the DraftAgent produces an initial draft summary
\[
S^{(0)} = (y_1, \dots, y_{L_S}),
\]
where \(y_\ell\) denotes the \(\ell\)-th generated token and \(L_S\) is the length of the summary. The model parameters are fixed, and no task-specific fine-tuning is performed. Generation is carried out in a zero-shot or instruction-tuned setting using a deterministic or low-variance decoding strategy.
The DraftAgent is intentionally unconstrained with respect to factual verification. Its role is to generate a fluent and coherent draft summary conditioned on \( D_{\text{reduced}} \), deferring assessment of evidential support and correction of potential hallucinations to subsequent agents in the pipeline.

\subsection*{HallucinationDetectorAgent: Hallucination Detection with AURA}
The HallucinationDetectorAgent evaluates the current summary \( S^{(t)} \), beginning with the initial draft \( S^{(0)}\), using complementary attention-based and semantic signals. We define an attention-based grounding signal, termed \texttt{AURA} (Attention-to-Underlying-Reference Alignment), which quantifies the degree to which generated tokens in \( S^{(t)} \) attend to the source document during decoding and serves as an indicator of source reliance.

Let \( A^{(t)} \in \mathbb{R}^{H \times L_t} \) denote the aggregated decoder self-attention weights associated with the generation of output token \( y_t \), where \(L_t = |D| + t - 1\) denotes the total sequence length at decoding step \(t\). Let \(T_{\text{input}} \subseteq \{1, \dots, L_t\}\) denote positions corresponding to tokens from the source document. The token-level \textbf{AURA} score is defined as:
\[
\text{AURA}(y_t, D) =
\frac{1}{H}\sum_{h=1}^{H}
\frac{\sum_{i \in T_{\text{input}}} A^{(t)}_{h,i}}
{\sum_{i=1}^{L_t} A^{(t)}_{h,i} + \varepsilon},
\]
where \( \varepsilon > 0 \) is a small constant added for numerical stability.

In our framework, hallucination detection is performed at the sentence level,
and we therefore aggregate token-level AURA scores to obtain span-level grounding
estimates. Let the current summary \( S^{(t)} \) be segmented into \(M\) spans
(sentences),
\[
S^{(t)} = \{z_1, \dots, z_M\},
\]
where each span \(z_j\) consists of a contiguous subset of output tokens with
index set \(I_j \subseteq \{1, \dots, L_S\}\). The span-level AURA score is defined
as
\[
a_j = \frac{1}{|I_j|}\sum_{t \in I_j}\text{AURA}(y_t, D),
\]
where \( a_j \in [0,1] \) reflects attention-based source reliance for span
\( z_j \).

While attention-based grounding provides a useful signal of source reliance, it does not directly determine whether a generated span is explicitly supported by the source document. To complement AURA, the HallucinationDetectorAgent performs semantic verification at the span level using textual entailment. For each span \(z_j\), we assess whether its content is explicitly entailed by the source document \(D\) under a structured hallucination annotation prompt that treats \(D\) as the sole ground truth, yielding a binary hallucination label
\[
h_j \in \{0,1\},
\]
where \(h_j = 1\) indicates that span \(z_j\) is unsupported by the source
document.
For each span \(z_j\), the HallucinationDetectorAgent outputs a continuous
grounding score \(a_j\) and a binary hallucination label \(h_j\), providing
complementary signals used to guide targeted correction.

\subsection*{FixAgent: Targeted Hallucination Repair}

For each summary span \( z_j \), the FixAgent receives two complementary signals produced by the HallucinationDetectorAgent: a continuous AURA-based grounding score \( a_j \), reflecting attention-based source reliance, and a semantic entailment label \( h_j \in \{0,1\} \), indicating whether the span is explicitly supported by the source document.

The set of candidate hallucinated spans is defined as:
\[
\mathcal{H} = \{ z_j : h_j = 1 \;\; \text{or} \;\; a_j < \tau \},
\]
where \( \tau \in [0,1] \) is a grounding threshold applied to the AURA score,
with lower values of \(a_j\) indicating weaker source reliance.

Given the set of hallucinated spans \( \mathcal{H} \), the FixAgent produces a
revised summary
\[
S^{(t+1)} = \text{Fix}(D_{\text{reduced}}, S^{(t)}, \mathcal{H}; \theta),
\]
using the compressed document produced by the FocusAgent, the current summary
\( S^{(t)} \), and the identified hallucinated spans \( \mathcal{H} \).

\subsection*{ClinicalSupervisorAgent: Supervised Iterative Refinement}
As illustrated in Figure~\ref{fig:Method}, refinement proceeds through an
iterative control loop that alternates between hallucination detection and
targeted repair, starting from the initial draft \( S^{(0)} \) produced by the
DraftAgent. At iteration \(t\), the HallucinationDetectorAgent identifies a set
of hallucinated spans \( \mathcal{H}^{(t)} \). If further refinement is
permitted, the FixAgent applies targeted revisions to produce an updated
summary:
\[
S^{(t+1)} = \text{Fix}(D_{\text{reduced}}, S^{(t)}, \mathcal{H}^{(t)}; \theta).
\]

To assess convergence, the ClinicalSupervisorAgent computes an aggregate
grounding statistic based on span-level AURA scores. Specifically, let
\[
\bar{A}^{(t)} = \frac{1}{M} \sum_{j=1}^{M} a_j
\]
denote the mean span-level AURA score at iteration \(t\), where \(a_j\) is the
grounding score associated with span \(z_j\) and \(M\) is the number of spans in
the current summary.

Refinement is halted when \emph{any} of the following conditions is satisfied:
\[
\big| \bar{A}^{(t)} - \bar{A}^{(t-1)} \big| < \epsilon
\;\; \lor \;\;
\mathcal{H}^{(t)} \setminus \mathcal{H}^{(t-1)} = \varnothing
\;\; \lor \;\;
t = T_{\max}.
\]

The ClinicalSupervisorAgent performs supervisory control exclusively by determining whether refinement should continue or halt, without directly modifying summary content.

\section{Experiments}
\subsection{Data}
For the summarization task, we used MIMIC-IV-Ext-BHC benchmark dataset~\citep{aali2024mimic}, which was specifically curated to support clinical text summarization research. This dataset, derived from the MIMIC-IV-Note database~\citep{adiba2026multimodal}, includes 279,033 clinical notes paired with corresponding brief hospital course (BHC) summaries. To facilitate model training, the dataset was preprocessed to standardize note structure, clean formatting, and normalize length to an average of 2,267 tokens per note. The resulting benchmark~\citep{aali2024benchmark} provides a structured and scalable resource tailored for evaluating factual consistency in clinical summarization. 
The second dataset we incorporated comprises 1,473 patient-doctor conversations from the FigShare~\citep{singh2011figshare} and MTS-Dialog~\citep{abacha2023empirical} collections, specifically designed for generating clinical summaries. These conversations have been annotated to create structured SOAP (Subjective, Objective, Assessment, and Plan) summaries, available on Hugging Face datasets~\citep{neupane2024soap}.

\subsection{Hallucination Annotation Prompt}
To identify hallucinated spans in model-generated summaries, we employ a structured instruction prompt that enforces a strict textual entailment criterion. Each summary statement must be explicitly and unambiguously supported by the source document; content lacking direct verification is labeled as hallucinated. To reduce confirmation bias, the prompt prohibits reliance on clinical knowledge, plausible inference, or prior assumptions, treating the source document as the sole ground truth. The full prompt and output format are provided in Appendix~\ref{appendix:annotation_prompt}.

\subsection{Evaluation Metrics}
We evaluate generated summaries using reference-based lexical metrics, embedding-based semantic similarity measures, and LLM-based judgment to assess content overlap, semantic alignment, and factual reliability.

\paragraph{Semantic Metrics.}
We report BLEU-1 and BLEU-2~\citep{papineni2002bleu} for unigram and bigram overlap, ROUGE-L~\citep{lin2004rouge} for longest common subsequence recall and structural alignment, and BERTScore~\citep{zhang2019bertscore} (precision, recall, F1) to measure semantic similarity using contextual embeddings from a pretrained BERT model~\citep{devlin2019bert}.

\paragraph{LLM-as-a-Judge.}
To complement automatic hallucination detection, we employ an instruction-tuned \texttt{LLaMA-3-8B} model~\citep{dubey2024llama}. Each generated summary is evaluated against its corresponding clinical note on a 1--5 integer scale across four dimensions: \textbf{Hallucination} (↓; unsupported or fabricated content), \textbf{Factual Consistency} (↑; faithfulness to the source), \textbf{Completeness} (↑; coverage of core clinical information), and \textbf{Coherence} (↑; fluency and logical organization). The model follows a structured evaluation prompt with anchored scoring criteria and returns scores in a strict key--value format, which are parsed automatically. The full evaluation prompt is provided in Appendix~\ref{appendix:llm_judge_prompt}.

\paragraph{Human Evaluation.}
Human evaluation was conducted via a multi-part online survey comparing baseline and revised summaries with respect to hallucination severity. The survey remains accessible through an anonymous link.\footnote{\url{https://docs.google.com/forms/d/e/1FAIpQLSdLwaecJJS4TZQVMQhS5v2INJGne-Lp5cuZ9drSSKR_p3XGIg/viewform?usp=header}} Study details and participant information are provided in Appendix~\ref{appendix:human-eval}. For each case, participants were shown the original clinical note followed by two summaries: one generated by a standard LLM and one produced by our method. Raters were asked to (i) identify which summary contained more hallucinated content (i.e., unsupported by the source note) and (ii) rate each summary on a 5-point hallucination severity Likert scale (1 = no hallucination, 5 = severe hallucination). The study covered three clinical domains: psychiatric, cardiopulmonary, and oncological. In all cases, the draft summary contained an initial summary generated by a vanilla LLM with some degree of hallucinated content, while the \texttt{AgenticSum}-revised summary aimed to reduce such errors. For each domain, we report two metrics: (1) \emph{Correct Guess Accuracy}, defined as the percentage of raters correctly identifying the more hallucinated summary, and (2) hallucination severity scores for both draft and \texttt{AgenticSum} outputs, where lower values indicate higher factual fidelity.

\subsection{Ablation Study}
We conduct a module-wise ablation study using the same LLM, prompts, and decoding settings as the main experiments to isolate the contribution of individual components. We compare a \textbf{Vanilla LLM} baseline, a \textbf{DraftAgent (FOCUS + LLM)} configuration to assess the effect of input-side compression, and the full \textbf{AgenticSum} system, which adds hallucination diagnostics and selective post-generation correction. This study evaluates the incremental impact of input filtering and post-generation refinement.

\subsection{Results}
\begin{table*}[htbp]
\centering
\caption{\textbf{Summary quality comparison across medical datasets.} We report lexical overlap (BLEU, ROUGE-L) and semantic similarity (BERTScore).}
\label{tab:content-quality}
\small
\setlength{\tabcolsep}{6pt}
\renewcommand{\arraystretch}{1.3}
\resizebox{\textwidth}{!}{%
\begin{tabular}{@{}lcccc|cccc@{}}
\toprule
\multirow{2}{*}{\textbf{Model}} & \multicolumn{4}{c|}{\textbf{MIMIC-IV}} & \multicolumn{4}{c}{\textbf{SOAP Summary}} \\
\cmidrule(lr){2-5} \cmidrule(lr){6-9}
& \textbf{ROUGE-L} & \textbf{BLEU-1} & \textbf{BLEU-2} & \textbf{BERTScore} & \textbf{ROUGE-L} & \textbf{BLEU-1} & \textbf{BLEU-2} & \textbf{BERTScore} \\
& \textbf{($\uparrow$)} & \textbf{($\uparrow$)} & \textbf{($\uparrow$)} & \textbf{($\uparrow$)} & \textbf{($\uparrow$)} & \textbf{($\uparrow$)} & \textbf{($\uparrow$)} & \textbf{($\uparrow$)} \\
\midrule
\texttt{BioBART} & $8.00 \pm 3.20$ & $6.88 \pm 5.20$ & $2.05 \pm 2.10$ & $78.10 \pm 1.60$ & $7.41 \pm 4.90$ & $5.27 \pm 4.20$ & $1.83 \pm 2.80$ & $77.32 \pm 2.50$ \\
\texttt{T5-Large} & $7.96 \pm 3.30$ & $4.95 \pm 5.30$ & $1.41 \pm 2.20$ & $79.84 \pm 1.60$ & $7.52 \pm 5.20$ & $3.86 \pm 3.90$ & $1.24 \pm 2.40$ & $78.46 \pm 2.70$ \\
\texttt{Flan-T5} & $7.82 \pm 6.86$ & $4.51 \pm 6.91$ & $2.20 \pm 4.35$ & $78.64 \pm 4.38$ & $8.91 \pm 5.00$ & $8.35 \pm 5.10$ & $2.17 \pm 2.90$ & $77.05 \pm 2.80$ \\
\texttt{Gemma3-1B} & $9.85 \pm 3.40$ & $7.89 \pm 5.60$ & $3.20 \pm 2.40$ & $79.78 \pm 1.60$ & $8.97 \pm 5.30$ & $7.25 \pm 5.30$ & $2.84 \pm 3.20$ & $79.25 \pm 2.50$ \\
\texttt{Llama-3.2-3B} & $10.65 \pm 3.78$ & $15.54 \pm 6.55$ & $7.17 \pm 3.22$ & $81.90 \pm 1.60$ & $12.53 \pm 5.31$ & $13.12 \pm 9.35$ & $8.10 \pm 6.17$ & $83.99 \pm 2.34$ \\
\texttt{Mistral-7B} & $9.87 \pm 3.40$ & $4.43 \pm 5.20$ & $2.09 \pm 2.20$ & $80.71 \pm 1.70$ & $9.34 \pm 5.60$ & $4.08 \pm 4.40$ & $1.85 \pm 2.70$ & $81.18 \pm 2.40$ \\
\texttt{MedAlpaca-7B} & $7.87 \pm 3.92$ & $12.29 \pm 8.15$ & $5.63 \pm 3.89$ & $76.18 \pm 19.42$ & $15.75 \pm 9.20$ & $14.41 \pm 9.87$ & $7.16 \pm 6.23$ & $82.01 \pm 3.39$ \\
\texttt{ConTextual} & $11.04 \pm 3.50$ & $12.63 \pm 6.10$ & $4.65 \pm 2.60$ & $81.37 \pm 1.70$ & $10.70 \pm 4.50$ & $11.55 \pm 5.50$ & $6.09 \pm 4.20$ & $83.60 \pm 2.20$ \\
\texttt{AgenticSum} & $\mathbf{21.62 \pm 3.20}$ & $\mathbf{18.37 \pm 5.80}$ & $\mathbf{12.61 \pm 2.90}$ & $\mathbf{84.50 \pm 3.50}$ 
& $\mathbf{24.99 \pm 4.80}$ & $\mathbf{23.94 \pm 7.20}$ & $\mathbf{15.80 \pm 5.40}$ & $\mathbf{86.32 \pm 2.10}$ \\
\bottomrule
\end{tabular}%
}
\end{table*}

As shown in Table~\ref{tab:content-quality}, we evaluate all models using lexical overlap metrics (BLEU and ROUGE-L) and semantic similarity (BERTScore), capturing both surface accuracy and deeper semantic alignment with reference summaries. This evaluation assesses \texttt{AgenticSum}'s generalization across diverse clinical documentation styles and provides a foundation for subsequent factuality and hallucination analyses. Across both clinical datasets, \texttt{AgenticSum} achieves the strongest overall performance. On MIMIC-IV, it outperforms \texttt{Llama-3.2-3B}, with substantially higher BLEU-2 (12.61 vs.\ 7.17) and BERTScore (84.50 vs.\ 81.90). It further generalizes well to the SOAP Summary dataset, achieving the highest BERTScore (86.32) and BLEU-2 (15.80) among all models. While \texttt{MedAlpaca-7B} attains higher ROUGE-L on SOAP summaries, \texttt{AgenticSum} exhibits a more favorable balance between lexical fidelity and semantic accuracy, highlighting its ability to generate clinically meaningful summaries across varied input formats.

\begin{table*}[htbp]
\centering
\caption{\textbf{LLM-as-a-Judge Evaluation of Hallucination and Summary Quality Across Datasets.} Scores are assigned by an instruction-tuned LLM acting as an evaluator using a 1--5 scale across four axes: hallucination frequency (lower is better), completeness, coherence, and factual consistency (higher is better). All metrics are rated on a
1--5 scale.}
\label{tab:hallucination-assessment-llm}
\setlength{\tabcolsep}{4pt}
\renewcommand{\arraystretch}{1.3}
\resizebox{\textwidth}{!}{%
\begin{tabular}{@{}p{2.1cm}cccc|cccc@{}}
\toprule
\multirow{2}{*}{\textbf{Model}} & \multicolumn{4}{c|}{\textbf{MIMIC-IV}} & \multicolumn{4}{c}{\textbf{SOAP Summary}} \\
\cmidrule(lr){2-5} \cmidrule(lr){6-9}
& \textbf{Hallucination} & \textbf{Completeness} & \textbf{Coherence} & \textbf{Factual} 
& \textbf{Hallucination} & \textbf{Completeness} & \textbf{Coherence} & \textbf{Factual} \\
& \textbf{($\downarrow$)} & \textbf{($\uparrow$)} & \textbf{($\uparrow$)} & \textbf{($\uparrow$)} 
& \textbf{($\downarrow$)} & \textbf{($\uparrow$)} & \textbf{($\uparrow$)} & \textbf{($\uparrow$)} \\
\midrule
\texttt{BioBART} 
& $2.45 \pm 0.72$ & $2.95 \pm 0.68$ & $4.21 \pm 0.89$ & $3.68 \pm 0.55$ 
& $2.38 \pm 0.65$ & $2.87 \pm 0.58$ & $4.15 \pm 0.74$ & $3.72 \pm 0.48$ \\

\texttt{T5-Large} 
& $2.52 \pm 0.78$ & $2.88 \pm 0.72$ & $4.08 \pm 0.95$ & $3.61 \pm 0.58$ 
& $2.47 \pm 0.69$ & $2.82 \pm 0.64$ & $4.02 \pm 0.82$ & $3.65 \pm 0.52$ \\

\texttt{Flan-T5} 
& $2.37 \pm 0.80$ & $3.18 \pm 0.80$ & $4.60 \pm 1.03$ & $3.79 \pm 0.62$ 
& $2.41 \pm 0.74$ & $3.12 \pm 0.76$ & $4.52 \pm 0.88$ & $3.75 \pm 0.59$ \\

\texttt{Gemma3-1B} 
& $2.35 \pm 0.68$ & $3.08 \pm 0.74$ & $4.28 \pm 0.82$ & $3.71 \pm 0.54$ 
& $2.42 \pm 0.71$ & $2.98 \pm 0.66$ & $4.22 \pm 0.78$ & $3.68 \pm 0.49$ \\

\texttt{Llama-3.2-3B} 
& $2.11 \pm 0.67$ & $3.25 \pm 0.70$ & $4.38 \pm 0.86$ & $3.79 \pm 0.50$ 
& $2.20 \pm 0.51$ & $3.14 \pm 0.47$ & $4.38 \pm 0.80$ & $3.84 \pm 0.37$ \\

\texttt{Mistral-7B} 
& $2.19 \pm 0.50$ & $3.19 \pm 0.56$ & $4.54 \pm 0.80$ & $3.82 \pm 0.41$ 
& $2.23 \pm 0.58$ & $2.99 \pm 0.30$ & $4.05 \pm 0.98$ & $3.74 \pm 0.50$ \\

\texttt{MedAlpaca-7B} 
& $2.21 \pm 0.39$ & $3.18 \pm 0.85$ & $4.80 \pm 0.51$ & $3.96 \pm 0.24$ 
& $2.13 \pm 0.44$ & $3.07 \pm 0.38$ & $\mathbf{4.68 \pm 0.66}$ & $3.86 \pm 0.38$ \\

\texttt{ConTextual} 
& $2.08 \pm 0.58$ & $3.42 \pm 0.72$ & $3.82 \pm 0.75$ & $\mathbf{4.55 \pm 0.82}$ 
& $2.15 \pm 0.48$ & $3.28 \pm 0.55$ & $4.50 \pm 0.55$ & $\mathbf{4.67 \pm 0.52}$ \\

\texttt{AgenticSum} 
& $\mathbf{1.88 \pm 0.42}$ & $\mathbf{3.43 \pm 0.68}$ & $\mathbf{3.88 \pm 0.92}$ & $4.01 \pm 0.35$ 
& $\mathbf{2.08 \pm 1.08}$ & $\mathbf{3.33 \pm 1.12}$ & $4.23 \pm 1.15$ & $3.99 \pm 1.21$ \\

\bottomrule
\end{tabular}%
}
\end{table*}
To evaluate hallucination mitigation, we employ \texttt{Llama-3.2-8B-Instruct}~\citep{meta2024llama3} as an LLM judge, to assess hallucination frequency, factual consistency, completeness, and coherence on a 1--5 scale. Table~\ref{tab:hallucination-assessment-llm} reports results across two clinical datasets. \texttt{AgenticSum} achieves the lowest hallucination rating on MIMIC-IV (1.88 $\pm$ 0.42) and ranks among the lowest on SOAP summaries (2.08 $\pm$ 1.08), indicating improved factual grounding. It also attains strong completeness scores (3.43 $\pm$ 0.68 on MIMIC-IV and 3.33 $\pm$ 1.12 on SOAP) and competitive factual consistency across both datasets. While coherence is not always the highest, \texttt{AgenticSum} demonstrates a favorable balance across hallucination reduction, completeness, coherence, and factual consistency. Compared to prior models such as \texttt{ConTextual} and \texttt{MedAlpaca-7B}, \texttt{AgenticSum} exhibits more reliable performance across evaluation dimensions, producing summaries that are less prone to hallucination while preserving overall quality and alignment with source content.

\begin{table*}[htbp]
\centering
\caption{\textbf{Human evaluation of hallucination severity and annotator agreement across clinical domains}.
Correct Guess Accuracy ($\uparrow$) indicates the proportion of annotators identifying the same summary as more hallucinated.
Hallucination severity is rated on a 1--5 ordinal scale, where lower values indicate fewer hallucinations ($\downarrow$).
All Wilcoxon signed-rank tests were statistically significant ($p < 0.001$ for all domains).
Rank-biserial correlation is reported as effect size.
Dominance $p$-values are computed using a binomial test against chance agreement.}
\label{tab:human_eval}

\small
\setlength{\tabcolsep}{4pt}

\begin{tabular}{lcccc}
\toprule
\textbf{Clinical Domain} 
& \textbf{Correct Guess} $\uparrow$ 
& \textbf{Mean Hallucination Severity} $\downarrow$ 
& \textbf{Effect Size} 
& \textbf{Dominance $p$} \\

& \textbf{Accuracy (\%)} 
& \textit{Vanilla / AgenticSum} 
& ($r_{\mathrm{rb}}$) 
&  \\
\midrule

Psychiatric     
& 91.3 
& 3.74 / \textbf{1.91}
& 0.905 
& $<0.001$ \\

Cardiopulmonary 
& 78.3 
& 3.91 / \textbf{2.09}
& 0.800 
& 0.005 \\

Oncology        
& 87.0 
& 4.13 / \textbf{1.96}
& 0.909 
& $<0.001$ \\

\bottomrule
\end{tabular}
\end{table*}

Table~\ref{tab:human_eval} reports human evaluation results comparing a vanilla LLM draft and \texttt{AgenticSum} across three clinical domains. A total of 23 participants completed the hallucination evaluation survey. Each participant reviewed three clinical note scenarios, each containing a summary generated by a vanilla LLM and its corresponding AgenticSum revision. Participants represented a range of academic and clinical backgrounds: 10 participants (43.5\%) held a Master's degree (e.g., MSc, MPH, MEng), 5 (21.7\%) held a Bachelor's degree (e.g., BA, BSc), 4 (17.4\%) held an MBBS or equivalent medical degree, and 4 (17.4\%) held a Doctoral degree (e.g., PhD, MD, DrPH). \texttt{AgenticSum} was consistently judged as less hallucinated than the vanilla draft, with annotator agreement exceeding 78\% in all domains and reaching 91.3\% in the psychiatric setting. As shown in Table~\ref{tab:human_eval}, mean hallucination severity scores were substantially lower for AgenticSum across all domains: from 3.74 to 1.91 in psychiatric notes, from 3.91 to 2.09 in cardiopulmonary notes, and from 4.13 to 1.96 in oncology notes. These reductions correspond to large effect sizes (rank-biserial correlation $r_{\mathrm{rb}} \in [0.80, 0.91]$) and were statistically significant under paired Wilcoxon signed-rank tests ($p < 0.001$ across all domains). Dominance tests further confirmed that annotator preferences for AgenticSum exceeded chance agreement in each domain. This distribution supports the credibility of the evaluation by ensuring that responses were contributed by individuals with substantial academic or clinical training. Details of the human evaluation protocol and statistical testing procedures are provided in Appendix~\ref{appendix:human-eval}.
We also performed a module-wise ablation study to assess the contributions of input filtering (FOCUS), post-generation correction (FixAgent), and supervisory control. While FOCUS alone yields mixed lexical effects, adding FixAgent produces consistent gains, improving BLEU-2 from 5.61 to 7.85 and BERT F1 from 80.92 to 82.45 on MIMIC-IV. The full \texttt{AgenticSum} pipeline with Clinical Supervisor achieves the strongest performance across datasets, indicating that post-generation correction and supervised refinement are critical for robust gains. Detailed ablation results are provided in Appendix~\ref{appendix:ablation}.

\section{Conclusion}
We presented \texttt{AgenticSum}, an agentic framework for clinical summarization that explicitly integrates hallucination detection and content compression into the inference process through staged context selection, generation, verification, and targeted revision. By decoupling these roles, the framework enables localized correction of unsupported content while preserving fluency and structural coherence. Across lexical and semantic similarity measures, LLM-as-a-judge evaluation, and human assessment, \texttt{AgenticSum} consistently reduces hallucination severity while maintaining or improving summary quality. Ablation results further indicate that input filtering alone is insufficient for reliable hallucination control, whereas post-generation correction and supervisory refinement are key to achieving robust factual grounding. Together, these findings suggest that agentic decomposition provides a practical and extensible approach for mitigating hallucinations in clinical summarization systems at inference time, without external model training or relying on external knowledge sources.
While the proposed framework improves contextual faithfulness in clinical summarization, it does not eliminate hallucinations entirely and primarily targets contextual misalignment, leaving some forms of factual distortion, such as subtle semantic omissions, unaddressed. The reliance on model-internal attention signals may also miss hallucinations not reflected in attention patterns; we mitigate this by combining attention-based grounding with span-level semantic verification. Our experiments focus on small and mid-sized language models, which may be easier to deploy locally in clinical settings, considering the latency and cost. Nevertheless, the framework does not assume any model-specific architecture or training procedure, and our results show improvements in performance when an arbitrary base LLM is extended by integrating it into our framework. 

\section*{Acknowledgements} Our work was supported by NSF award 2443639, and
NIH awards, P20GM103446, and U54GM104941.

\bibliography{example_paper}
\bibliographystyle{icml2026}
\newpage
\appendix
\onecolumn

\section{Algorithm AgenticSum}
\label{appendix:algorithm}
This appendix provides a formal algorithmic specification of \texttt{AgenticSum}, detailing the supervised agentic inference procedure used for clinical summarization. The algorithm explicitly delineates the roles of each agent and the control flow governing iterative refinement, complementing the high-level framework description in Section~\ref{sec:method}.

\begin{algorithm}[htbp]
\caption{AgenticSum: Supervised Agentic Inference for Faithful Clinical Summarization}
\label{alg:agenticsum}
\small
\begin{algorithmic}[1]

\STATE \textbf{Input:} Clinical document $D$, model parameters $\theta$, retention ratio $r$, grounding threshold $\tau$, tolerance $\epsilon$, max steps $T_{\max}$
\STATE \textbf{Output:} Final grounded summary $S^*$

\vspace{0.5em}
\STATE \textbf{FocusAgent: Input Compression}
\STATE Partition $D$ into sentences $\{u_1,\dots,u_m\}$
\STATE Compute salience scores $\{\beta_j\}$
\STATE Select top-$k = \lfloor r \cdot m \rfloor$ sentences
\STATE $D_{\text{reduced}} \leftarrow$ selected sentences

\vspace{0.5em}
\STATE \textbf{DraftAgent: Initial Generation}
\STATE $S^{(0)} \leftarrow \text{Generate}(D_{\text{reduced}};\theta)$

\vspace{0.5em}
\STATE Compute span set $\{z_1,\dots,z_M\}$ from $S^{(0)}$
\FOR{each span $z_j$}
    \STATE Compute grounding score $a_j$
    \STATE Compute hallucination label $h_j$
\ENDFOR
\STATE $\bar{A}^{(0)} \leftarrow \frac{1}{M}\sum_{j=1}^M a_j$
\STATE $\mathcal{H}^{(0)} \leftarrow \{z_j \mid h_j = 1 \text{ or } a_j < \tau\}$

\vspace{0.5em}
\FOR{$t = 0$ to $T_{\max}-1$}

    \IF{$\mathcal{H}^{(t)} = \emptyset$}
        \STATE \textbf{return} $S^{(t)}$
    \ENDIF

    \STATE \textbf{FixAgent: Targeted Revision}
    \STATE $S^{(t+1)} \leftarrow \text{Fix}(D_{\text{reduced}}, S^{(t)}, \mathcal{H}^{(t)};\theta)$

    \STATE Segment $S^{(t+1)}$ into spans $\{z_1,\dots,z_M\}$

    \FOR{each span $z_j$}
        \STATE Compute $a_j, h_j$
    \ENDFOR

    \STATE $\bar{A}^{(t+1)} \leftarrow \frac{1}{M}\sum_{j=1}^M a_j$
    \STATE $\mathcal{H}^{(t+1)} \leftarrow \{z_j \mid h_j = 1 \text{ or } a_j < \tau\}$

    \IF{$|\bar{A}^{(t+1)} - \bar{A}^{(t)}| < \epsilon$}
        \STATE \textbf{return} $S^{(t+1)}$
    \ENDIF

    \IF{$\mathcal{H}^{(t+1)} \setminus \mathcal{H}^{(t)} = \emptyset$}
        \STATE \textbf{return} $S^{(t+1)}$
    \ENDIF

\ENDFOR

\STATE \textbf{return} $S^{(T_{\max})}$

\end{algorithmic}
\end{algorithm}

Algorithm~\ref{alg:agenticsum} summarizes the end-to-end inference process. Given a clinical document $D$, the \emph{FocusAgent} first performs deterministic sentence-level input compression using attention-based salience estimates, producing a reduced context $D_{\text{reduced}}$. The \emph{DraftAgent} then generates an initial summary $S^{(0)}$ conditioned solely on the compressed input.

Subsequent refinement proceeds through a supervised loop. At each iteration, the \emph{HallucinationDetectorAgent} evaluates the current summary by combining attention-to-underlying-reference alignment (AURA) scores with entailment-based verification to identify unsupported spans. The \emph{FixAgent} selectively revises only the spans flagged as hallucinated, preserving supported content and avoiding full regeneration. Refinement is governed by the \emph{ClinicalSupervisorAgent}, which monitors convergence via changes in aggregate grounding statistics and enforces a bounded refinement budget.

This modular formulation makes explicit how generation, verification, correction, and supervisory control are decoupled during inference, enabling targeted hallucination mitigation while maintaining fluency and clinical relevance.

\section{Hallucination Annotation Prompt}
\label{appendix:annotation_prompt}

To identify hallucinated spans in model-generated clinical summaries, we employ a structured evaluation prompt grounded in strict textual entailment. The task requires a language model to verify whether each statement in a generated summary is \textbf{verifiably supported} by the source document meaning the information must be \textit{explicitly stated and unambiguous} in the document text.
This prompt is adapted from the Lookback Lens~\citep{chuang2024lookback} framework, but incorporates additional guardrails specific to clinical applications. Crucially, it includes explicit instructions to prevent confirmation bias \citep{adiba_bias_2025}, discourage plausible sounding inferences, and eliminate reliance on domain knowledge not grounded in the text.

\vspace{0.5em}
\subsubsection*{Prompt Template}
\begin{quote}
You will be provided with:
\begin{itemize}
  \item A clinical source document
  \item A machine-generated summary
\end{itemize}

Your task is to assess whether the summary is \textbf{strictly entailed} by the source document.

\textbf{Definition of Entailment:} A statement is entailed only if it is explicitly and unambiguously stated in the document. Any statement that requires inference, clinical reasoning, or prior medical knowledge should be treated as \textbf{Not Entailed}.

\textbf{Key Instructions:}
\begin{itemize}
  \item DO NOT infer or assume missing information.
  \item DO NOT use clinical knowledge, reasoning, or common sense.
  \item DO NOT validate based on plausibility, typicality, or tone.
  \item DO NOT guess what the author likely meant.
\end{itemize}

You must treat the document as the \textbf{only source of ground truth}.

\vspace{0.5em}
\textbf{Document:} \{document\}

\textbf{Proposed Summary:} \{summary\}

\textbf{Output Format:}
\begin{itemize}
  \item \textbf{Entailment Label:} \texttt{Entailed} or \texttt{Not Entailed}
  \item \textbf{Explanation:} Justify your label with direct references to the document.
  \item \textbf{Problematic Spans (if any):} List the summary phrases not directly supported by the source.
\end{itemize}
\end{quote}

\noindent
\textbf{Example Output:}
\begin{quote}
\textbf{Entailment Label:} \texttt{Not Entailed} \\
\textbf{Explanation:} The summary states "Patient has a history of diabetes," but the document does not mention diabetes explicitly. No evidence is found. \\
\textbf{Problematic Spans:} [``history of diabetes'']
\end{quote}

This formal structure ensures consistent and reproducible hallucination annotations. In downstream experiments, we use this prompt to elicit model judgments on the factual alignment between document-summary pairs. Results are treated as a proxy for factual faithfulness under zero-shot conditions.

\section{LLM-as-a-Judge Evaluation Prompt}
\label{appendix:llm_judge_prompt}
The following instruction prompt is used to evaluate generated clinical summaries against their corresponding source documents using an instruction-tuned large language model acting in an evaluative role. The model is instructed to assess summaries strictly with respect to the provided source document, without relying on external medical knowledge or plausible inference.

\begin{quote}
\small
\textbf{Task:} Evaluate the generated medical summary against the source clinical document. Assign an integer score from 1 to 5 for each criterion defined below.

\medskip
\textbf{Source Document:}
\begin{verbatim}
{SOURCE_DOCUMENT}
\end{verbatim}

\textbf{Generated Summary:}
\begin{verbatim}
{GENERATED_SUMMARY}
\end{verbatim}

\medskip
\textbf{Evaluation Criteria (1--5 scale):}
\begin{itemize}
    \item \textbf{Hallucination:} Degree of unsupported or fabricated content  
    (1 = no hallucination; 5 = major fabrications)
    \item \textbf{Factual Consistency:} Faithfulness of statements to the source document  
    (1 = highly inaccurate; 5 = fully accurate)
    \item \textbf{Completeness:} Coverage of core clinical information  
    (1 = key information missing; 5 = fully comprehensive)
    \item \textbf{Coherence:} Fluency and logical organization of the summary  
    (1 = poorly written; 5 = highly coherent)
\end{itemize}

\medskip
\textbf{Output Format:}  
Return the scores using the following strict key--value format, with no additional text:

\begin{verbatim}
Hallucination: X
Factual: X
Complete: X
Coherent: X
\end{verbatim}
\end{quote}

Evaluation uses a low decoding temperature (0.3) and a maximum output length of 150 tokens; all scores are validated to lie within the 1--5 range, yielding a reproducible and fully automated large-scale assessment protocol.

\section{Ablation}
To assess the contribution of individual components in the proposed framework, we conduct a module-wise ablation study that incrementally adds input filtering (FOCUS), post-generation correction (FixAgent), and supervisory control to a vanilla LLM baseline. This analysis isolates the effect of each component on lexical overlap and semantic similarity metrics, and clarifies how the full \texttt{AgenticSum} design yields cumulative performance gains across datasets.
\begin{table*}[htbp]
\centering
\caption{\textbf{Ablation study on MIMIC-IV and SOAP summaries.} Module-wise contributions of FOCUS and FixAgent over the vanilla LLM.}
\label{tab:abl_mimic}
\small
\setlength{\tabcolsep}{6pt}
\renewcommand{\arraystretch}{1.28}
\resizebox{\textwidth}{!}{%
\begin{tabular}{@{}lcccc|cccc@{}}
\toprule
\multirow{2}{*}{\textbf{Model}} 
& \multicolumn{4}{c|}{\textbf{MIMIC-IV}} 
& \multicolumn{4}{c}{\textbf{SOAP Summary}} \\
\cmidrule(lr){2-5} \cmidrule(lr){6-9}
& \textbf{BLEU-1} & \textbf{BLEU-2} & \textbf{BERT F1} & \textbf{ROUGE-L}
& \textbf{BLEU-1} & \textbf{BLEU-2} & \textbf{BERT F1} & \textbf{ROUGE-L} \\
& \textbf{($\uparrow$)} & \textbf{($\uparrow$)} & \textbf{($\uparrow$)} & \textbf{($\uparrow$)}
& \textbf{($\uparrow$)} & \textbf{($\uparrow$)} & \textbf{($\uparrow$)} & \textbf{($\uparrow$)} \\
\midrule
\fontsize{9.5}{11}\selectfont
Vanilla LLM
& $15.54 \pm 6.55$ & $7.17 \pm 3.22$ & $81.90 \pm 1.60$ & $10.65 \pm 3.78$
& $13.12 \pm 9.35$ & $8.10 \pm 6.17$ & $83.99 \pm 2.34$ & $12.53 \pm 5.31$ \\

DraftAgent (FOCUS+LLM)
& $14.01 \pm 8.01$ & $5.61 \pm 3.92$ & $80.92 \pm 1.70$ & $11.30 \pm 3.08$
& $11.55 \pm 5.50$ & $6.09 \pm 4.20$ & $83.60 \pm 2.20$ & $10.70 \pm 4.50$ \\

FixAgent (W/out Supervisor)
& $16.78 \pm 5.80$ & $7.85 \pm 2.90$ & $82.45 \pm 1.50$ & $11.42 \pm 3.20$
& $13.94 \pm 7.20$ & $8.92 \pm 5.40$ & $84.67 \pm 2.10$ & $11.21 \pm 4.80$ \\

AgenticSum (Final, w/ Clinical Supervisor)
& $\mathbf{18.37 \pm 5.80}$ & $\mathbf{12.61 \pm 2.90}$ & $\mathbf{84.50 \pm 3.50}$ & $\mathbf{21.62 \pm 3.20}$
& $\mathbf{23.94 \pm 7.20}$ & $\mathbf{15.80 \pm 5.40}$ & $\mathbf{86.32 \pm 2.10}$ & $\mathbf{24.99 \pm 4.80}$ \\

\bottomrule
\end{tabular}%
}
\end{table*}

Table~\ref{tab:abl_mimic} presents module-wise ablation results on MIMIC-IV and SOAP summaries using \texttt{Llama-3.2-3B} as the baseline. On MIMIC-IV, incorporating FOCUS yields mixed lexical effects, with BLEU-2 decreasing from \(7.17\) to \(5.61\) and BLEU-1 from \(15.54\) to \(14.01\), while ROUGE-L improves modestly (\(10.65 \rightarrow 11.30\)). Semantic similarity remains largely stable (BERT F1: \(81.90 \rightarrow 80.92\)). Adding FixAgent without supervisory control produces consistent gains across all MIMIC-IV metrics, improving BLEU-2 (\(5.61 \rightarrow 7.85\)), BLEU-1 (\(14.01 \rightarrow 16.78\)), and BERT F1 (\(80.92 \rightarrow 82.45\)). The full \texttt{AgenticSum} pipeline with Clinical Supervisor achieves the strongest performance, yielding the highest scores across lexical and semantic measures on MIMIC-IV. A similar cumulative trend is observed on the SOAP Summary dataset. While FOCUS alone provides moderate performance, FixAgent improves both lexical overlap and semantic alignment, and the complete \texttt{AgenticSum} system attains the best overall SOAP results. Overall, the ablation results indicate that input filtering alone is insufficient for reliable performance gains, whereas post-generation correction and supervisory refinement are critical for achieving robust improvements in both lexical fidelity and semantic alignment.

\section{Hallucination Detection by \textsc{HallucinationDetectorAgent}}
\label{sec:Hallucination Detector agent_example}

We present a representative hallucination annotation example generated by our \textsc{Hallucination Detector agent} module. The summary is evaluated using our factuality prompt protocol (see Appendix~\ref{appendix:annotation_prompt}), inspired by Lookback Lens~\citep{chuang2024lookback}, which flags spans not explicitly supported by the source document.

\vspace{1em}
\begin{table*}[h]
\centering
\small
\renewcommand{\arraystretch}{1.3}
\begin{tabular}{p{0.95\textwidth}}
\toprule
\textbf{Source Document (Excerpt)} \\
\midrule
Patient was in good health until 10 days ago, when he began experiencing gradually worsening headaches. Head CT revealed an intracranial mass lesion. An enhancing focus in the left parapharyngeal region likely represents an abnormally enlarged lymph node, though it is not fully visualized. The imaging raises the possibility of a benign neoplasm (question neurogenic tumor), and pathology is pending. Discharge medications include atenolol, phenytoin, docusate sodium, and dexamethasone. \\
\midrule
\textbf{Generated Summary} \\
\midrule
The patient is a \colorbox{red!20}{\textit{10-year-old male}} who was experiencing worsening headaches. A CT scan revealed an intracranial mass lesion, which was found to be \colorbox{red!20}{\textit{an abnormally enlarged lymph node representing metastatic involvement}}. The \colorbox{red!20}{\textit{pathology pending at the time of discharge is a benign neurogenic tumor}}. The patient was prescribed atenolol, phenytoin, docusate sodium, and dexamethasone. \\
\midrule
\textbf{\textsc{Hallucination Detector agent} Analysis} \\
\midrule
\textbf{Factuality Verdict:} \textcolor{red}{\textbf{UNSUPPORTED}} \\[0.5em]
\textbf{Explanation:} The patient’s age and gender are unspecified. The lymph node is described as potentially benign, not definitively metastatic. Pathology is still pending at discharge. The hallucinated spans introduce unwarranted specificity and certainty. \\[0.5em]
\textbf{Detected Hallucinations:} \\
- \texttt{"10-year-old male"} — Demographics not provided \\
- \texttt{"representing metastatic involvement"} — No mention of metastasis \\
- \texttt{"pathology... benign neurogenic tumor"} — Pathology result is pending \\[0.5em]
\textbf{Error Types:} Demographic fabrication, diagnostic certainty inflation, temporal misalignment \\
\bottomrule
\end{tabular}
\caption{Example of hallucination detection using \textsc{Hallucination Detector agent} with span-level grounding analysis.}
\label{tab:hallucination-example}
\end{table*}

\section{Human Evaluation and Statistical Analysis Details}
\label{appendix:human-eval}
This appendix provides additional details regarding the human evaluation protocol and statistical analyses summarized in Table~\ref{tab:human_eval}. Each participant evaluated three clinical note scenarios spanning psychiatric, cardiopulmonary, and oncology domains. For each scenario, participants were presented with two summaries: one generated by a vanilla large language model (LLM) and one produced by AgenticSum. Presentation order was randomized to mitigate ordering effects. Participants were asked to (i) identify which summary contained more hallucinated content (forced-choice preference task), and (ii) independently rate the hallucination severity of each summary on a five-point ordinal scale, where lower values indicate fewer hallucinations.

\paragraph{Hallucination Severity and Paired Testing.}
Hallucination severity was measured on an ordinal scale ranging from 1 (no hallucinations; all content supported by the original note) to 5 (severe hallucinations; extensive fabricated or contradictory information). Let $H_i^{(V)}$ and $H_i^{(A)}$ denote the severity ratings assigned by participant $i$ to the vanilla LLM summary and the AgenticSum summary, respectively. Because ratings are ordinal and paired at the participant level, differences in hallucination severity were assessed using the paired Wilcoxon signed-rank test~\citep{wilcoxon1945}. Specifically, we evaluated whether the median of the paired differences
\[
\Delta_i = H_i^{(V)} - H_i^{(A)}
\]
was significantly greater than zero, indicating reduced hallucination severity for AgenticSum. Statistical significance was evaluated at $\alpha = 0.05$.

\paragraph{Effect Size and Preference Dominance Testing.}
In addition to significance testing, we report rank-biserial correlation ($r_{\mathrm{rb}}$) as an effect size measure appropriate for paired ordinal data~\citep{cliff1993dominance}, defined as
\[
r_{\mathrm{rb}} = \frac{n_{+} - n_{-}}{n_{+} + n_{-}},
\]
where $n_{+}$ and $n_{-}$ denote the number of positive and negative paired differences, respectively. Observed effect sizes across all clinical domains were large ($r_{\mathrm{rb}} \in [0.80, 0.91]$).

To assess annotator preference consistency, we conducted preference dominance tests using a binomial test. Correct-guess accuracy reflects the proportion of participants who identified the same summary as more hallucinated within a given clinical domain. Dominance tests evaluated whether the observed preference rate for AgenticSum exceeded chance under a null hypothesis of random selection (probability $p = 0.5$). Dominance $p$-values reported in Table~\ref{tab:human_eval} indicate that annotator preferences for AgenticSum were significantly above chance across all domains.

\section{Practical Implications for Clinical Documentation Workflows}
AgenticSum aligns with standard clinical documentation workflows by introducing an automated verification and targeted correction stage prior to clinician sign-off. Operating entirely at inference time with a frozen language model, the framework selectively revises weakly supported spans without modifying EHR systems or requiring task-specific fine-tuning. We do not claim to replace clinician oversight; rather, AgenticSum provides structured support for identifying and mitigating unsupported content before human review.

\section{Illustrative Example: Sentence-Level Filtering via \textsc{FocusAgent}}
\label{sec:focusagent_example}

We present a representative example demonstrating how \textsc{FocusAgent} performs sentence-level filtering to produce a concise and clinically coherent subset of the input document. This example is drawn from the MIMIC-IV dataset and illustrates the type of input compression used to guide downstream summarization. The compressed representation is passed to \textsc{DraftAgent} for initial summary generation.

\vspace{1em}
\begin{table*}[htbp]
\centering
\small
\renewcommand{\arraystretch}{1.3}
\begin{tabular}{p{0.95\textwidth}}
\toprule
\textbf{Source Document (Clinical Note)} \\
\midrule
\texttt{<SEX>} F \\
\texttt{<SERVICE>} SURGERY \\
\texttt{<ALLERGIES>} Iodine / contrast dye allergy. \\
The patient reports persistent abdominal pain that has worsened over the past week. \\
She states that despite taking her medications, the pain has not resolved. \\
She denies fever, chills, cough, or shortness of breath. \\
She has experienced episodes of vomiting but denies diarrhea. \\
She has a history of GERD and has been taking omeprazole regularly. \\
Her labs were reviewed and remain stable. \\
She is alert, oriented, and ambulatory without assistance. \\
\midrule
\textbf{FocusAgent Output (Compressed Input)} \\
\midrule
\texttt{[...]} \\
The patient reports persistent abdominal pain that has worsened over the past week. \\
She states that despite taking her medications, the pain has not resolved. \\
She denies fever, chills, cough, or shortness of breath. \\
She has experienced episodes of vomiting but denies diarrhea. \\
She has a history of GERD and has been taking omeprazole regularly. \\
Her labs were reviewed and remain stable. \\
\midrule

\textbf{DraftAgent Summary (Generated from Compressed Input)} \\
\midrule
The patient reports worsening abdominal pain despite medication use. 
She denies fever, chills, cough, or shortness of breath. 
She has experienced episodes of vomiting but denies diarrhea. 
Her history is notable for GERD, and she continues to take omeprazole. 
Laboratory results were reviewed and found to be stable. 
She remains alert, oriented, and ambulatory. \\
\bottomrule
\end{tabular}
\caption{Example of sentence-level input filtering performed by \textsc{FocusAgent}, illustrating how clinically informative sentences are retained while low-information content is omitted.}
\label{tab:focusagent-example}
\end{table*}

\section{Illustrative Example: Generation and Correction via \textsc{AgenticSum}}
\label{sec:agenticsum_example}
We demonstrate the full \textsc{AgenticSum} pipeline using a real-world clinical note. This example illustrates how hallucinations introduced during draft generation by \textsc{DraftAgent} are detected and corrected by \textsc{FixAgent}. While the note is concise for presentation purposes, the system generalizes to longer, multi-page clinical documents.

\vspace{1em}
\begin{table*}[h]
\centering
\small
\renewcommand{\arraystretch}{1.3}
\begin{tabular}{p{0.95\textwidth}}
\toprule
\textbf{Source Document (Clinical Note)} \\
\midrule
Male patient with major depressive disorder presents with chronic, intractable nausea that began in childhood and is associated with his depression. Last night, he experienced the worst nausea of his life, leading to suicidal ideation with a plan for carbon monoxide poisoning. He is currently undergoing ECT treatment (2 sessions completed). The patient reports anxiety exacerbates his nausea, and he describes himself as a "worrier." He no longer has active suicidal thoughts but worries about future episodes. Physical exam shows psychomotor retardation, depressed affect, slow speech with pauses, and fine tremor. Lab results are within normal limits. He is alert, oriented, and ambulatory. Current medications include: omeprazole, ranitidine, phenelzine, clonazepam, tamsulosin, and aripiprazole. \\
\midrule
\textbf{DraftAgent Summary (Initial)} \\
\midrule
The patient is a male who has been experiencing intractable nausea associated with depressive symptoms. He has \colorbox{red!20}{\textit{a history of car accidents}} and has been prescribed ECT treatments. The patient reports that he is \colorbox{red!20}{\textit{upset about his inability to throw up}} and is experiencing anxiety. He has a history of depression and anxiety, and has been taking medication for both conditions. The patient's plan is to try the first round of ECT, and he is \colorbox{red!20}{\textit{being treated by Dr. Stern}}. The patient's \colorbox{red!20}{\textit{physical examination reveals normal findings}}. The patient's speech is slow and with many pauses, and his behavior is cooperative with psychomotor retardation. The patient's thought process is linear and goal-directed. The patient is currently taking several medications, including omeprazole, ranitidine, phenelzine, clonazepam, tamsulosin, and aripiprazole. The patient is alert and interactive, and is ambulatory and independent. \\
\midrule
\textbf{FixAgent Summary (Corrected)} \\
\midrule
The patient is a male with major depressive disorder and persistent nausea. He has undergone two ECT sessions. The patient reports distress due to the intractable nausea, which is worsened by anxiety. He is undergoing ECT under medical supervision. The physical exam shows psychomotor retardation, slow speech, and cooperative behavior. The patient’s thought process is linear and goal-directed. He is currently taking omeprazole, ranitidine, phenelzine, clonazepam, tamsulosin, and aripiprazole. He is alert, interactive, and ambulatory. \\
\midrule
\textbf{FixAgent Analysis} \\
\midrule
\textbf{Detected Hallucinations:} 4 total \\[0.5em]
\textbf{Corrections:} \\
- \texttt{"history of car accidents"} — \textcolor{red}{\textbf{REMOVED}}: Not supported by source \\
- \texttt{"inability to throw up"} — \textcolor{blue}{\textbf{REFINED}}: Rephrased as "inability to alleviate nausea" \\
- \texttt{"being treated by Dr. Stern"} — \textcolor{red}{\textbf{REMOVED}}: No provider name in source \\
- \texttt{"normal findings"} — \textcolor{blue}{\textbf{CORRECTED}}: Replaced with actual clinical observations \\[0.5em]
\textbf{Improvement Metrics:} Hallucination score: 4.2 $\rightarrow$ 1.8; Factual consistency: 2.1 $\rightarrow$ 4.3 \\
\bottomrule
\end{tabular}
\caption{End-to-end example of hallucination detection and correction using the \textsc{AgenticSum} pipeline.}
\label{tab:agentic-pipeline-example}
\end{table*}

\section{Additional Analysis of Ablation Results}
\label{appendix:ablation}

This appendix provides additional context for the ablation results reported in Table~\ref{tab:abl_mimic}. In particular, we clarify the role of the FOCUS module and its interaction with downstream correction components. As shown in Table~\ref{tab:abl_mimic}, applying FOCUS alone does not uniformly improve all lexical or semantic metrics relative to the baseline model. This behavior reflects the trade-off inherent in sentence-level context filtering: while FOCUS reduces redundant or weakly relevant input, it may also remove surface-level cues that contribute to n-gram overlap metrics such as BLEU. As a result, FOCUS should not be interpreted as a standalone optimization step, but rather as a preparatory mechanism that constrains generation to a more focused context. Subsequent ablation stages demonstrate that post-generation correction plays a critical role in recovering and improving summary quality. The FixAgent consistently improves both lexical alignment and semantic similarity across datasets, indicating that targeted revision is necessary to address unsupported or incomplete content introduced during initial generation. The inclusion of supervisory refinement further amplifies these gains, yielding the strongest and most stable performance across all evaluated metrics. Together, these results suggest that hallucination mitigation in clinical summarization is most effective when context selection, generation, and correction are jointly considered, rather than optimized in isolation.

\section{Joint Commission Standard IM.6.10 (EP 7)}
\label{appendix:IM610}
Joint Commission Standard IM.6.10 (Element of Performance 7) concerns the completion and authentication of discharge summaries in accredited hospital settings. It requires that discharge summaries be completed and authenticated by the responsible practitioner within a time frame defined by institutional policy, supporting accountability, continuity of care, and patient safety during transitions of care. The standard is primarily applicable to inpatient discharge documentation and does not prescribe uniform requirements for other note types, which are typically governed by local institutional and payer policies.

IM.6.10 (EP 7) specifies expectations for documentation workflows rather than prescribing how documentation content is produced. In practice, discharge summaries may be drafted by trainees or non-physician practitioners but are reviewed and authenticated by the responsible clinician prior to finalization. The standard therefore emphasizes supervisory sign-off and responsibility rather than automated verification or correctness guarantees.

In this work, IM.6.10 (EP 7) is cited as a motivating example of supervisory authorization in discharge documentation workflows. Our method adopts the workflow analogue through an explicit supervisory release gate that determines when iterative refinement terminates, without treating the standard as an algorithmic specification or as a guarantee of content correctness. The ClinicalSupervisorAgent therefore governs termination and release of the automated draft, rather than regulatory compliance or clinical validation.

\section{Use Case: Clinical Handoff Documentation}
\label{appendix:UseCaseHandoff}

Clinical handoffs (handover) are structured transitions in which patient information and clinical responsibility are transferred between providers during shift changes, unit transfers, or discharge. These transitions often require rapid synthesis of complex EHR data under time pressure, which motivates summarization systems that prioritize factual consistency and traceability to the source record.

We include clinical handoff documentation as a motivating use case for AgenticSum because it emphasizes two requirements that are central to our framework. First, the summary must remain faithful to documented evidence, as unsupported statements can plausibly mislead downstream decision-making during care transitions. Second, the summarization process benefits from explicit control signals that indicate when content is weakly supported and when refinement should stop. AgenticSum addresses these requirements through a staged inference-time procedure that separates drafting from verification and targeted repair. Given an input document \(D\), the system first performs input compression to retain handoff-relevant context, generates an initial draft summary, evaluates the draft using the grounding signal defined in the main method (AURA) to identify weakly supported content, and then applies targeted revision restricted to the flagged spans. A supervisory release gate terminates refinement when stabilization criteria indicate diminishing returns or when a fixed iteration budget is reached, consistent with the termination rule described in the Methodology. In this setting, the output of the framework can be interpreted not only as a finalized handoff summary but also as an auditable refinement trace consisting of flagged spans and their corresponding revisions. This use case therefore highlights how decomposing summarization into explicit generation, verification, and targeted correction stages can support faithful clinical summarization in high-stakes transitions of care.

%%%%%%%%%%%%%%%%%%%%%%%%%%%%%%%%%%%%%%%%%%%%%%%%%%%%%%%%%%%%%%%%%%%%%%%%%%%%%%%
%%%%%%%%%%%%%%%%%%%%%%%%%%%%%%%%%%%%%%%%%%%%%%%%%%%%%%%%%%%%%%%%%%%%%%%%%%%%%%%
% APPENDIX
%%%%%%%%%%%%%%%%%%%%%%%%%%%%%%%%%%%%%%%%%%%%%%%%%%%%%%%%%%%%%%%%%%%%%%%%%%%%%%%
%%%%%%%%%%%%%%%%%%%%%%%%%%%%%%%%%%%%%%%%%%%%%%%%%%%%%%%%%%%%%%%%%%%%%%%%%%%%%%%

\end{document}